\documentclass[journal]{IEEEtran}
\usepackage{amsmath,amsfonts}
\usepackage{algorithmic}
\usepackage{algorithm}
\usepackage{array}
\usepackage[caption=false,font=normalsize,labelfont=sf,textfont=sf]{subfig}
\usepackage{textcomp}
\usepackage{stfloats}
\usepackage{url}
\usepackage{verbatim}
\usepackage{graphicx}
\usepackage{cite}
\usepackage{orcidlink}
\usepackage{tabu}

\begin{document}

\title{Under the Cover Infant Pose Estimation\\ using Multimodal Data}

\author{Daniel G. Kyrollos \orcidlink{0000-0002-1203-959X}, Anthony Fuller \orcidlink{0000-0001-8187-5850}, Kim Greenwood \orcidlink{0000-0002-1300-4866}, JoAnn Harrold \orcidlink{0000-0002-3218-6570}, James R. Green \orcidlink{0000-0002-6039-2355}
\thanks{This research is funded by the Natural Sciences and Engineering Research Council of Canada and the IBM Center for Advanced Studies.}
\thanks{Daniel G. Kyrollos, Anthony Fuller and James R. Green are with Carleton University Biomedical Informatics Collaboratory, Department of Systems and Computer Engineering, Carleton University, Ottawa, Canada. Email: daniel.kyrollos@carleton.ca}
\thanks{Kim Greenwood is with the Department of Clinical Engineering, Children’s Hospital of Eastern Ontario, Ottawa, Canada}
\thanks{JoAnn Harrold is with the Department of Neonatology, Children’s Hospital of Eastern Ontario, Ottawa, Canada}
}

\markboth{Journal of \LaTeX\ Class Files,~Vol.~14, No.~8, August~2021}%
{Shell \MakeLowercase{\textit{et al.}}: A Sample Article Using IEEEtran.cls for IEEE Journals}


\maketitle

\begin{abstract}
Infant pose monitoring during sleep has multiple applications in both healthcare and home settings. In a healthcare setting, pose detection can be used for region of interest detection and movement detection for noncontact based monitoring systems. In a home setting, pose detection can be used to detect sleep   positions which has shown to have a strong influence on multiple health factors. However, pose monitoring during sleep is challenging due to heavy occlusions from blanket coverings and low lighting. To address this, we present a novel dataset, \textbf{S}imultaneously-collected multimodal \textbf{Ma}nnequin \textbf{L}ying pose (SMaL) dataset, for under the cover infant pose estimation. We collect depth and pressure imagery of an infant mannequin in different poses under various cover conditions. We successfully infer full body pose under the cover by training state-of-art pose estimation methods and leveraging existing multimodal adult pose datasets for transfer learning. We demonstrate a hierarchical pretraining strategy for transformer-based models to significantly improve performance on our dataset. Our best performing model was able to detect joints under the cover within 25mm 86\% of the time with an overall mean error of 16.9mm. Data, code and models publicly available at https://github.com/DanielKyr/SMaL
\end{abstract}


\section{Introduction}
\IEEEPARstart{I}{n} the neonatal intensive care unit (NICU), continuous monitoring is required due to the neonate's fragile health condition. This often involves the use of wired sensors placed on the chest area and limbs to measure and track a newborn’s physiologic signals. However, false alarms can be generated due to motion artifacts created at the skin-electrode interface during patient movement \cite{Hamilton2000}. Studies have shown that 87.5\% of alarms in the NICU are false alarms, with a majority caused by movements of the infants \cite{Wolf1996}.In addition, the adhesives used for the electrodes can irritate the fragile neonate’s skin \cite{adhesive1997}. The present study is part of a larger research initiative that aims to address these issues by developing novel unobtrusive neonatal patient monitoring systems based on pressure sensitive mats (PSMs) and colour/depth (RGB-D) video for the purpose of monitoring vital signs and detecting patient movements for the reduction of false alarms.

Initial studies showed that it is possible to detect patient movement in order to gate false alarms arising from motion artifacts \cite{Kyrollos2021-saspsm}. These preliminary results showed that general motion detection was insufficient to to create a robust system and that more precises movement tracking was required. Precisely detecting the neonate's individual limb motion through pose monitoring would allow a system to better associate false alarms caused on specific sensors with the movement of the limb on which the sensor was placed. Pose monitoring can also be beneficial to non-contact vital sign monitoring techniques, which rely on region of interest (ROI) detection \cite{Maurya2021}. One can extract relevant anatomical regions from pose, such as the chest, which then can be used for respiration rate (RR) estimation using color, depth, or pressure based technologies \cite{Kyrollos2021-sasrr, Nizami2018}. In addition, since these methods are sensitive to motion artifacts \cite{Maurya2021}, pose detection could also be used to reject estimates where the patient is moving significantly. Outside of a healthcare setting, monitoring an infant's pose during sleep can provide critical information about an infant's sleeping position. The sleeping position of an infant has a strong influence on multiple health factors, such as sudden infant death syndrome (SIDS) \cite{aap1992positioning}, cerebral oxygenation \cite{Bembich2012}, heart rate variability (HRV) \cite{Fister2020}, and obstructive sleep apnea \cite{Kukkola2022}. 

It is clear that pose monitoring is a critical part of NICU and home monitoring systems. However, standard pose estimation techniques are not suited for monitoring during sleep since the environment introduces adverse vision conditions: heavy occlusions due to blanket coverings and low lighting (including complete darkness for premature infants in isolettes). These challenging conditions can be mitigating by employing additional sensing modalities other than color imagery, such as contact pressure and depth data. To partially address this challenge, Liu \textit{et al.} released an in-bed human pose dataset, called \textbf{S}imultaneously-collected multimodal \textbf{L}ying \textbf{P}ose (SLP) \cite{SLPLiu2022}. The SLP dataset contains adult pose data collected under three conditions: no cover, a thin sheet, and a thick blanket, using four sensing modalities: color imaging, thermal imaging, depth sensing, and pressure sensing. However, this dataset only contains adult subjects and cannot be directly applied to infants. Therefore, there is a need for an equivalent dataset specifically for infants. Collecting an equivalent dataset using real infants would not be feasible since infants will not hold their position between cover conditions; instead, a poseable infant mannequin can be used. With this in mind, we introduce the \textbf{S}imultaneously-collected multimodal \textbf{Ma}nnequin \textbf{L}ying pose (SMaL) dataset. This dataset contains a set of 300 unique poses under three cover conditions using three sensor modalities: color imaging, depth sensing, and pressure sensing. Since we are leveraging a mannequin, thermal imaging is not relevant. The SMaL dataset represents the first multimodal dataset for infant pose estimation and the first dataset to explore under the cover pose estimation for infants.  

The SMaL dataset is a relatively small dataset when compared to the SLP dataset. This represents a common challenge in the infant pose estimation field - when compared to adult datasets, infant data is limited. A common approach to address this small data problem is the use of transfer learning. The two common sources of additional data are adult datasets and synthetic datasets \cite{Huang2021}. To improve performance on our SMaL dataset we leverage two additional data sources for transfer learning. The first is the SLP dataset, which contains labelled adult depth and PSM images \cite{SLPLiu2022}, which we use for additional training data for pose estimation. The second is BodyPressureSD which contains simulated depth and pressure images of adults in lying positions under blanket covering \cite{Clever2022}. With the recent surge in vision transformer models and their superior performance, we utilize these data as a source for a self-supervised pretraining task for a vision transformer pose estimation model. We employ the methods of Reed \textit{et al.} for hierarchical pretraining and propose our own hierarchical pretraining strategy for infant pose estimation \cite{reed2022self}. To the best of knowledge, this work represents the \textit{first} study which employs hierarchical pretraining for infant pose estimation.

This work extends a previous work titled \textit{"Transfer Learning Approaches for Neonate Head Localization from Pressure Images"} \cite{Kyrollos2022}, which demonstrated: 1) successful transfer of annotations across domains (\textit{i.e.}, RGB to PSM), and 2) improved performance on the head detection task when using a CNN backbone previously trained on adult pressure imagery. The present study extends these approaches in order to estimate full body pose, with 14 joints, including the head, from a combination of depth and pressure imagery. We successfully trained pose estimation models to infer under the cover infant pose from depth and pressure imagery. We also demonstrate that our hierarchical pretraining strategy significantly improves accuracy on the SMaL dataset. The main contributions of the current works are summarized as follows:

\begin{enumerate}
    \item Present a mannequin infant in-bed pose dataset, SMaL, with simultaneously-collected multimodal imagery: RGB, depth and pressure under different cover conditions.  
    \item Train and compare various state of the art pose estimation models on SMaL while leveraging adult data for transfer learning 
    \item  Establish the benefit of hierarchical pretraining for transformer-based pose estimation models when applied to multimodal data
\end{enumerate}

Section \ref{works} covers related literature. Section \ref{data} presents the data collection methodology for the SMaL dataset. Section \ref{vits} describes the vision transformer architecture and the hierarchical pretraining paradigm. Section \ref{training} presents the pose estimation model architectures and the training strategies used for each ,as well as how we evaluate the methods. This is followed by the results and discussion in Section \ref{results}.

\section{Related Works}\label{works}
\textbf{{In-bed Adult Pose Estimation.}}
Although there is a multitude of works addressing human pose estimation, only few address the resting condition where the human is is lying in a bed. This is firstly due to the fact that, until recently, there was a lack of available large-scale datasets. Secondly, although RGB data has been leveraged for at-rest posture estimation, these methods are constrained to well illuminated environments and to cases with little-to-no occlusion \cite{Liu2017}. To overcome these challenges, researchers have leveraged additional modalities for this task. Pressure data has been used extensively for estimation of 2D pose \cite{Davoodnia2021}, 3D pose \cite{Clever2018,Casas2018}, and even complete 3D human pose and body shape \cite{Clever2020}. However, when solely using pressure images for pose estimation, ambiguities arise from free-moving limbs producing negligible contact pressure \cite{Clever2018, Kyrollos2021-i2mtc}. Depth data has also been leveraged for estimating human pose during sleep. Clever \textit{et al.} were able to estimate full 3D adult pose and body shape and estimate a corresponding pressure image under the cover using depth imagery \cite{Clever2022}. To train their model, they generated synthetic data via a soft-body physics simulation of a human body, a mattress, a pressure sensing mat, and a covering blanket. Their publicly available dataset, BodyPressureSD contains approximately 100k simulated depth and pressure images of adults in lying positions under blanket covering. Currently, the SLP dataset is the benchmark dataset for in-bed pose estimation \cite{SLPLiu2022}. It is a large-scale dataset with over 100 subjects and nearly 15,000 pose images, which is comparable to other well-known general purpose human pose datasets \cite{SLPLiu2022}. The posture coverage evenly comprises supine, left, and right side sleep posture categories. Each subject was asked to perform 15 poses for each category under 3 cover conditions using 4 imaging modalities simultaneously. The size of the dataset and its comprehensive set of modalities makes the SLP dataset versatile for different vision task, exemplified by the recent works that leverage this dataset \cite{Cao2022human,Dayarathna2022,Yin2022,Afham2022}.

\textbf{Infant Pose Estimation.}
There is a significant amount of research focused on infant pose estimation. The main motivation behind developing these methods is to create automatic systems that can perform general movements assessment (GMA) for diagnosis of cerebral palsy and other developmental disabilities. RGB and depth are the two main modalities utilized for this task. There are various works that investigate RGB images for infant pose detection describing a wide range of infant pose estimation methods, including both CNN-based \cite{Groos2022} and transformer-based \cite{Cao2022} methods. Since there are not many public infant datasets when compared to adult datasets, some works opt to use synthetic data for training \cite{Hesse2019,Huang2021}. 

Rather than RGB images, several studies have opted for depth imagery, due to its privacy preserving nature. Moccia \textit{et al.} released a dataset of 16k annotated depth images from the 27 patients in the NICU for infant pose estimation, called \textit{babyPose} \cite{Moccia2020}. They also trained a CNN using spatio-temporal features on consecutive depth frames. Wu \textit{et al.} curated a dataset of 50k annotated depth images from the 27 patients in the NICU. They labelled this data semi-automatically by running existing pose estimators on the RGB images and transferring the annotations to the depth stream. To our knowledge, no work has attempted infant pose estimation from contact pressure imagery; however, some works have explored posture recognition \cite{Donati2013,Rihar2014}. Our previous work, explored head localization from pressure imagery but not the full pose \cite{Kyrollos2022}. 

Although, as mentioned, there are depth-based infant pose estimation datasets \cite{Moccia2020,Wu2022} these have focused on applications for GMA and do not consider scenarios in which the infant is fully occluded with coverings. In addition, no pressure-based or multimodal data exists for infant pose estimation. In this work, we aim to fill these gaps through our mannequin infant dataset, SMaL, which includes simultaneously-collected RGB, depth and pressure under different cover conditions.  This is the \textit{first} dataset to explore lying infant pose using multiple modalities and under heavy occlusions. 

\section{SMaL Dataset}\label{data}
\subsection{Experimental Set-Up}
\begin{figure}[htbp]
\centerline{\includegraphics[width=\hsize]{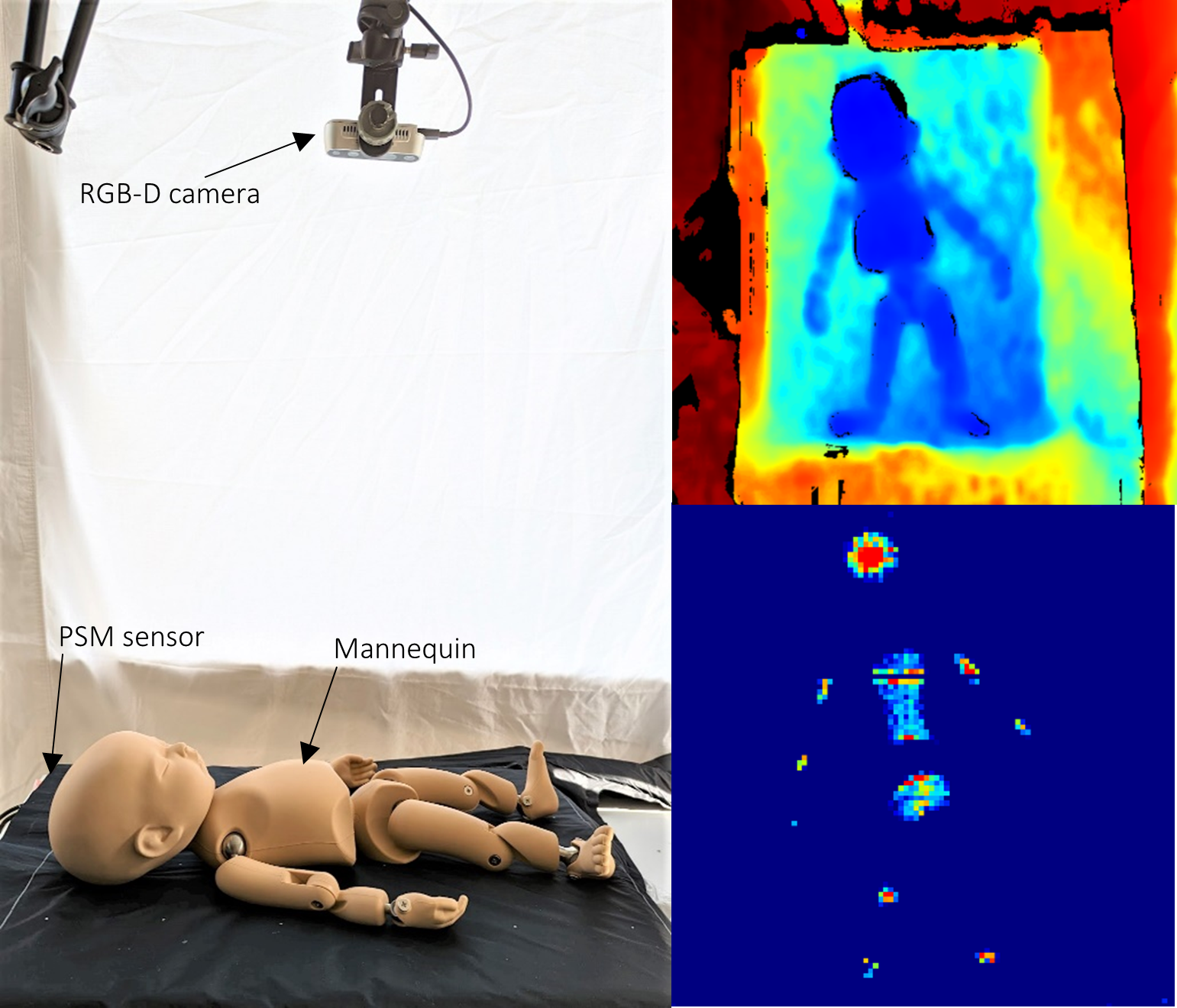}}
\caption{Experimental setup for data collection (left); the PSM is between the mattress and the mannequin and the RGB-D camera is overhead. Sample data from the depth stream (top right) and the PSM (bottom right) are shown.}
\label{fig:set-up}
\end{figure}

The data collection set-up includes an Intel RealSense D435 camera to capture colour and depth video data (© Intel Corporation, Santa Clara, CA, USA) and a LX100:100.100.05 PSM (XSensor Technology Corp,
Calgary, AB, Canada). The D435 camera has a resolution of up to 1280 × 720 px for RGB and 1920 × 1080 px for depth. It uses stereo vision to calculate depth and an infrared projector to project a non-visible static IR pattern to improve depth accuracy in scenes with low texture. The depth accuracy is $\pm2\%$ at 2 meters The PSM sensor has a spatial resolution of 5.08 mm with an overall sensing area of 50.8 × 50.8 cm$^2$, made up of a grid of 100 × 100 sensels. It captures ballistographic signals ranging from 0.0 to 3.41 N/cm$^2$. The camera was placed overhead of the PSM at a distance of 0.5 meters with a small foam mattress underneath. The mannequin used was the StandInBaby® mannequin (StandInBaby, Springfield Lakes Qld., Australia). It features articulated joints that cover a full range of natural motion, accurate size and weight distribution representing the 50\textsuperscript{th} percentile of a newborn infant (50cm length, 6.8lbs weight). The photo of the experimental set-up is shown in Figure \ref{fig:set-up}, with sample data from the depth and PSM streams.

\subsection{Multimodal Registration}\label{registration}
A grid of visible landmarks were drawn on the PSM in order to obtain reliable registration between the RGB and PSM streams, as demonstrated in \cite{Kyrollos2021-i2mtc}. The landmarks form a 6-by-6 grid with a 9 cm separation between each point. The true locations of these points are known in the PSM plane by their position relative to the border marked by the manufacturer. Since these points are visible by the RGB-D video camera, their location in the video plane is precisely known. Before any data collection session, the location of the visible landmarks were identified in the RGB stream. With the known location of these landmarks in the PSM stream, a homography was calculated using a RANSAC-based robust method. The depth stream was aligned to the PSM by first converting the RGB coordinates of the landmarks to depth coordinates using the camera parameters, then calculating a new homography matrix for the depth stream.

\subsection{Data Collection}
\begin{figure*}[ht]
\centerline{\includegraphics[width=\hsize]{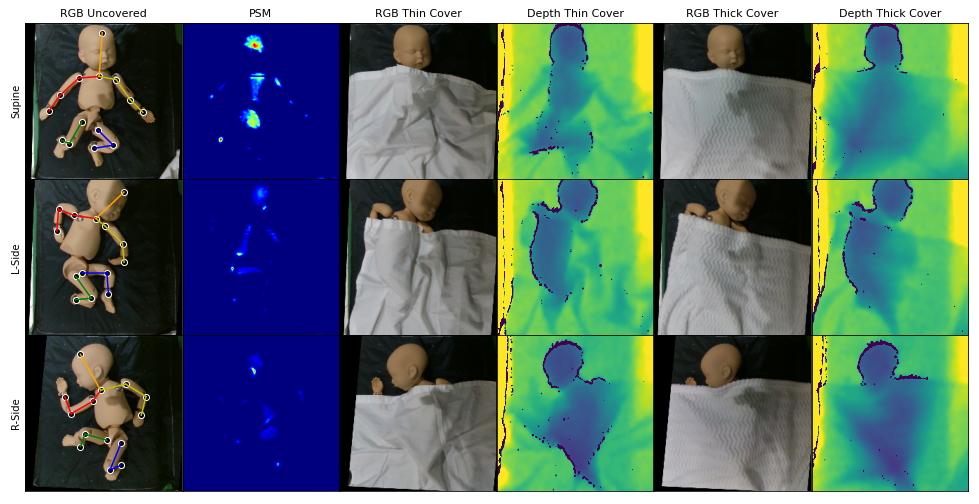}}
\caption{Examples from the SMaL dataset. RGB, depth, and PSM are shown under  under all three coverage conditions for the three pose categories}
\label{fig:example}
\end{figure*}

The infant manikin pose data were collected using a custom software with a GUI. The software live-streamed the RGB and depth data and was used to annotate registration points as well as pose annotation during data collection. The PSM data were recorded using the XSensor Pro 8 software and was time-aligned to the camera data following data collection. For each pose, the mannequin was placed in a novel position, the pose was annotated on the RGB stream and an image was captured for each of three cover conditions: uncovered, thin cover, and thick cover. The thin cover was a bed sheet that was approximately 1mm in thickness, while the thick cover was a blanket that was approximately 5mm in thickness. The pose annotation followed that of the SLP dataset, focusing on the major limbs for a total of 14 joints: Ankle-R, Knee-R, Hip-R, Hip-L, Knee-L, Ankle-L, Wrist-R, Elbow-R, Shoulder-R, Shoulder-L, Elbow-L, Wrist-L, Thorax, and Head Top, where L and R stand for the left and right side, respectively. Poses were categorized into three categories: supine position, left side, and right side, to match the SLP dataset. For each category, 100 poses were captured for a total of 300 unique poses each under the three cover conditions. To transfer the annotations to the PSM from RGB, the homography matrix calculated from the RGB-PSM registration was used. This annotation was transferred to all the cover conditions. The depth image was then aligned using the depth to PSM homography. The final data contained the depth aligned image, the pressure image and the pose annotation in PSM coordinates. The final images were sized to 224 × 224 to be compatible with the vision transformer backbones. Figure \ref{fig:example} displays some examples from the dataset. 

\section{Vision Transformers}\label{vits}
\subsection{Pretraining and Fine-tuning}
Pretraining refers to training a model on a pretext task using a large quantity of data, often in the millions of samples, in order to learn general knowledge stored in the form of learned parameters. Fine-tuning refers to transferring this general knowledge to application-specific downstream tasks. This knowledge transfer is accomplished by initializing model parameters with the parameters learned through pretraining, then adjusting them via supervised learning on annotated data. Pretraining can be accomplished through supervised learning, which trains models using labels gathered through annotation; pretraining can also be accomplished through self-supervised learning (SSL), which trains models using labels generated through an SSL algorithm. For example, labels can be generated by hiding portions of input data from the model, then training the model to reconstruct the hidden content. This is a popular class of SSL algorithms often referred to as reconstructive learning and is recently exemplified by masked autoencoding (MAE \cite{he2022masked})—which we leverage in section \ref{training}. A model architecture that fits incredibly well in the “pretraining and fine-tuning” paradigm is the transformer. 

\subsection{Transformer Architecture}
Introduced by Vaswani et al. \cite{vaswani2017attention}, the transformer is an attention-based model architecture used across many applications. When a “vanilla” transformer is used in computer vision applications, it is called a vision transformer (ViT) \cite{dosovitskiy2020image}. After adapting pretrained ViTs to specific tasks, ViTs achieve state-of-the-art performance in image classification \cite{he2022masked,peng2022beit}, semantic segmentation \cite{peng2022beit}, object detection \cite{li2022exploring}, and, importantly for our work, pose estimation \cite{vitpose}.

ViTs first process an image into a set of “patches”; then, these patches are processed by a transformer model, which outputs patch encodings. Processing an image into a set of patches is accomplished by splitting the image into small squares, usually 16 by 16 pixels, resulting in patch tensors of shape 16 × 16 × 3. Since our images have a height and width of 224 pixels, dividing by 16 pixels per patch leaves us with a 14 by 14 grid of patches, which are then arranged in a set of 196 total patches. Next, these patch tensors are flattened, resulting in 768-d feature vectors, still retaining the original pixel values. In the penultimate patchification step, each of the 768-d feature vectors is linearly projected to the transformer model width. In this work, we use ViT-Base models, which have a model width of 768. Thus, this linear projection outputs 768-d feature vectors that we call patch representations. In the final patchification step, each patch representation gets added to a unique sinusoidal position embedding that represents the location of each patch within the image. Now, each patch representation contains both content and positional information.

Next, this set of patch representations is processed by a transformer model composed of stacked transformer blocks. Each transformer block is composed of a self-attention sublayer and a feedforward sublayer. Self-attention sublayers learn to exchange features between patches, and feedforward sublayers learn to transform the features of individual patches. The ViT-Base architecture employs 12 transformer blocks to encode each patch with contextual “higher level” features. The term “ViT encoder” refers to both patchification and transformer modules.

\subsection{Masked Autoencoding (MAE)}
MAE is an SSL algorithm that pretrains ViTs by encoding a small portion of patches (referred to as visible patches), then employs a ViT decoder to reconstruct the hidden patches. In more detail, MAE first encodes a randomly sampled 25\% of image patches using a ViT encoder and holds the remaining 75\% of patches aside. The ViT decoder receives visible patch encodings and mask embeddings. These embeddings represent the hidden patches and can be thought of as placeholders for the decoder to write. Before the decoder processes these inputs, another set of sinusoidal position embeddings are added to both visible patch encodings and mask embeddings; these position embeddings represent the locations of patches within the original image. Given these inputs, the decoder outputs an image prediction, matching the image’s shape of 224 × 224 × 3. The loss is then computed by the mean squared error between the original held-out patches and the decoder’s predictions of the held-out patches. The decoder’s predictions of visible patches are ignored. Both encoder and decoder are trained end-to-end; after pretraining, the decoder is typically discarded, and only the ViT encoder is leveraged through fine-tuning on downstream tasks. When pretraining with MAE, a smaller decoder size is typically chosen to reduce the computational costs of pretraining; specifically, they chose a decoder model width of 512 and 2 transformer blocks. This asymmetric encoder-decoder design reduces pretraining FLOPs by approximately 3x \cite{he2022masked}. The baseline ViT encoder we use in this article was pretrained by \textit{He at al.} \cite{he2022masked} for 1600 epochs on the ImageNet-1k dataset. 

\subsection{Hierarchical Pretraining}
The similarity between upstream (pretraining) and downstream (fine-tuning) tasks is a leading indicator of downstream performance \cite{abnar2021exploring}. In our case, we do not have access to a model pretrained on fused depth and pressure data. Consequently, any pretrained model we initialize our model with will result in dissimilar upstream and downstream tasks; this is not ideal for knowledge transfer. Facing the same challenge, Reed and Yue et al. \cite{reed2022self} propose hierarchical pretraining, which progressively pretrains models on data closer to their downstream data, starting from a model pretrained on ImageNet. We leverage hierarchical pretraining by continuing to pretrain our ViT encoder on in-domain (i.e., fused depth and pressure) data using the MAE SSL algorithm. This results in three pretrained models that have been specialized to our domain by pretraining on (i) in-domain simulated data, (ii) in-domain real data, and (iii) first in-domain simulated, then in-domain real data. We compare these three models to our baseline model, which did not undergo further in-domain pretraining. For our first two specialized models, we pretrain on in-domain—either simulated or real—data for 150 epochs, initializing our ViT encoder from MAE’s pretrained model parameters and our ViT decoder randomly. Our decoder must be initialized randomly because the MAE authors do not share their decoder model parameters. For our third specialized model, we first pretrain on in-domain simulated data for 150 epochs, initializing our ViT encoder from MAE’s pretrained model parameters and our ViT decoder randomly. Then, we continue to pretrain on in-domain real data for 150 epochs, initializing our ViT encoder and decoder model parameters from the prior stage. This leaves us with three specialized ViT encoders, which we leverage through fine-tuning on annotated pose estimation data.

\section{Under the Cover Infant Pose Estimation}\label{training}
We train and test state-of-art pose estimation methods on our SMaL dataset. We compare various model architectures and various approaches to the hierarchical pretraining for the pure vision transformer models.

\subsection{Datasets}
\textbf{BodyPressureSD}
BodyPressureSD contains 97,495 unique body shapes, poses, and image samples. Each sample consists of of a human model with a realistic lying pose. From this model, simulated depth images in covered and uncovered settings and a corresponding physics-based pressure image were simulated.  The depth image shape is 128 × 54 and the pressure image shape is 64 × 27. They are spatially aligned and can be stacked. Both uncovered and covered conditions were used, for a total dataset size of 194,990 images.

\begin{figure}[htbp]
\centerline{\includegraphics[scale=0.55]{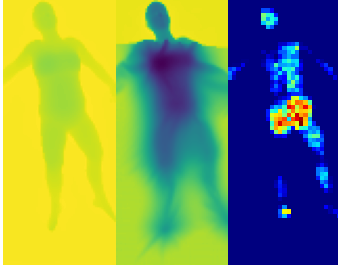}}
\caption{A sample from BodyPressureSD, showing uncovered depth, covered depth and pressure.}
\label{fig:synth}
\end{figure} 

\textbf{SLP}
SLP contains 4590 real adult in-bed poses. For each sample, there were three cover conditions (\textit{i.e.} uncovered, thin cover, thick cover). The depth image shape is 512 × 424 and the pressure image shape is 192 × 84. The images were aligned using the provided homography. The dataset was split into a training set of 4050 and the remaining was used as a validation set. All cover conditions were used for training for a total of 12,150 images; only covered conditions were used for validation for a total of 1080 images.

\begin{figure}[htbp]
\centerline{\includegraphics[width=\hsize]{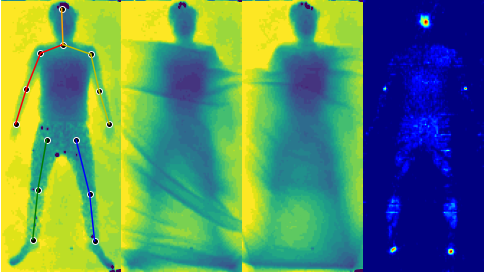}}
\caption{A sample from SLP, showing all three cover conditions in depth and corresponding pressure image. Joint labels are used for training and are equivalent to SMaL joints.}
\label{fig:slp}
\end{figure} 

\textbf{SMaL}
SMaL contains 300 mannequin poses. For each sample, there were three cover conditions (\textit{i.e.} uncovered, thin cover, thick cover). The depth image size is 640 × 480 and the pressure image shape is 100 × 100. The images were aligned using registration process described in Section \ref{registration}. The dataset was split into five folds of 60 each for cross-validation, due to the small size of the dataset. Three folds were used for training, one as validation for early stopping and one for testing. All cover conditions were used for training for a total of 540 images; only covered conditions were used for validation and test for a total of 120 images for each.

\subsection{Training and Evaluation}
The effect of pretraining and transfer learning on the models' accuracy were investigated. The synthetic BodyPressureSD dataset was used solely for pretraining the ViT backbone. The SLP dataset was used for both pretraining the ViT backbone and as the initial training dataset for transfer learning to the SMaL dataset. The input image for all the models had shape with 224 × 224 with two channels, one for depth and one for pressure. The depth channel was duplicated to be compatible with pretrained ImageNet backbones.    

\subsubsection{Training Settings}
Finetuning the models was done from scratch and comprised of two stages. The first stage was finetuning on the training set of SLP and using the validation for early stopping. The second stage was continuing the finetuning using the SMaL training set with early stopping on the validation set and a final evaluation on the test set. The second stage of finetuning was repeated for the 5-fold cross-validation. The joint locations were transformed to heatmaps by setting joint location to the maximum point of a 2-dimensional Gaussian distribution, with sigma equal to 1. The heatmap method has become a standard for human pose estimation and is employed by most of the state of the art methods. These heatmaps have a shape of 56 × 56 to match the downsampling factor of 4 that is commonly used. Therefore the final targets had a shape of 56 × 56 × 14, one channel for each joint. The loss function used was mean squared error, with the AdamW optimizer, 5 warmup epochs and half-cycle cosine learning rate decay. Major augmentations were applied when training SLP to prevent overfitting. Only slight rotation and horizontal flipping were used during SMaL training as these are the only variations that exist in the dataset given the limited dataset size. A summary of the training settings for the SLP and SMaL datasets are shown in Table \ref{tab:hyper}.

\begin{table}[ht]
\centering
\caption{Training Parameters}
\label{tab:hyper}
\begin{tabular}{c|cc}
\hline
\textbf{Parameters} &  \textbf{SLP} & \textbf{SMaL}\\
\hline         
Batch Size & 256 & 16\\
Learning Rate & 1e-3 & 1e-4\\
Training Iterations & 50 & 50\\
Warm Up Iterations & 5 & 5\\
\hline
\textbf{Augmentations} &  \textbf{SLP} & \textbf{SMaL}\\
\hline
Rotation Factor & $<$30$^{\circ}$ & $<$2$^{\circ}$ \\
Scaling Factor & $<$0.25 & 0 \\
Do Occlusion & 50\% & 0 \\
Color Scaling &  $<$0.2 & 0 \\
Do H-Flip & 50\% & 50\%\\
\hline
\end{tabular}
\end{table}

\subsubsection{Evaluation}
Two metrics were used to evaluate the models on the SMaL dataset: percent of correct keypoints (PCK) and the normalized mean error (NME). To count as a correct keypoint for PCK, the predicted keypoint had to be within 0.05 of the normalized distance from the ground truth keypoint. A normalized distance threshold can be used because all images are at the same scale after applying the camera to PSM homography. Since the real scale of the PSM is known, a 0.05 threshold is equivalent to 25.4mm (each sensel corresponds to 5.08 mm). Equivalently, the NME is  multiplied by the real world scale to measure true error in mm. Only the PCK is used for early stopping on the validation set during both stages of finetuning. Both metrics are used when evaluating on the test set.

\subsection{Model Architectures}
\subsubsection{HRNet}
HighResolution Net (HRNet) is a purely CNN based approach to pose estimation.HRNet is able to maintain high resolution representations throughout the whole neural network, opposed to previous methods which recover high resolution representations from low resolution representations. This is possible through the parallel nature of HRNet which exchanges the information across parallel multi-resolution subnetworks over and over and finally recovers joint estimates from the high resolution network. There are two HRNet variants,  one small net and one big net: HRNet-W32 and HRNet-W48, where 32 and 48 represent the widths of the high-resolution subnetworks in the last three stages, respectively.  \cite{hrnet}

\subsubsection{TransPose}
TransPose model consists of three components: a CNN backbone to extract low-level image features, a transformer model to capture long-range spatial interactions between feature vectors across locations and a head to predict the keypoints heatmaps. Only the initial several layers of the original ImageNet pretrained CNNs are used to extract features from images, the parameters numbers of which are only about 5.5\% and 25\%, for ResNet and HRNet respectively. After passing the image through a CNN the feature maps are flattened and used as input to the transformer which arranges its output into the 2D-structure heatmaps. The variants of Transpose include different backbones (\textit{i.e.}, TransPose-R has a ResNet and TransPose-H has HRNet) and different number of transformer blocks (\textit{i.e.}, TransPose-H-A4 has 4 blocks, TransPose-H-A6 has 6). TransPose outperformed both HRNet variants on the COCO dataset. \cite{transpose}

\subsubsection{ViTPose}
ViTPose employs a pretrained ViT encoder and a lightweight decoder for pose estimation. The ViT encoder produces a 768-d representation of each patch. After arranging the encoder’s outputs in a 14 by 14 grid of patches, the entire image is represented by a tensor of shape 14 × 14 × 768. This tensor is then processed by a decoder that upsamples the tensor via two deconvolutional layers resulting in a pose prediction tensor of shape 56 × 56 × 14, where 14 is the number of channels (one for each joint). Before fine-tuning ViTPose on annotated pose estimation data, the ViTPose encoder is initialized from a pretrained ViT encoder. In ViTPose, they initialize with MAE’s encoder; we use this as a baseline to compare with ViTPose models initialized with our three specialized encoders. The base ViTPose outperformed TransPose and HRNet on the COC0 dataset. \cite{vitpose}

\begin{figure*}[ht]
\centerline{\includegraphics[width=\hsize]{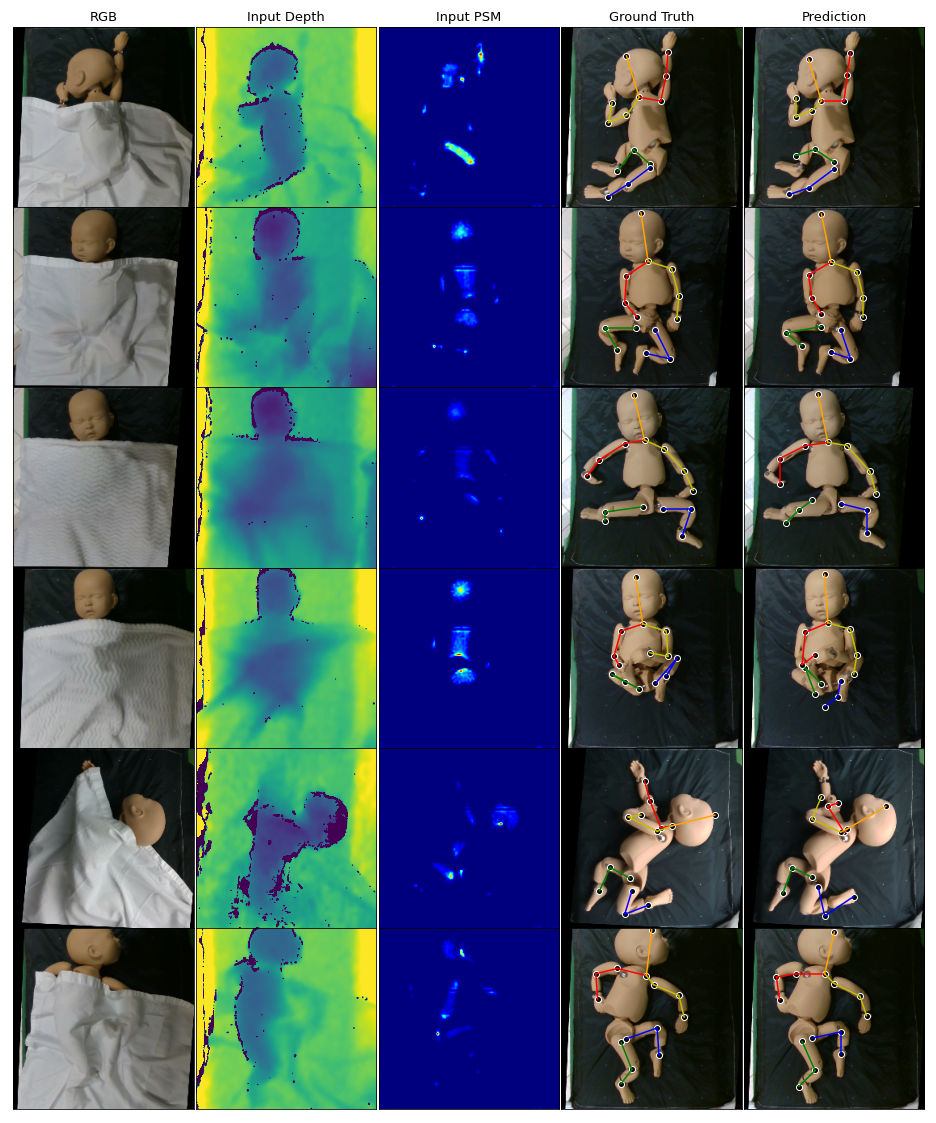}}
\caption{Predictions on a test set from the best performing model, ViTPose-S-150, along with ground truth data.}
\label{fig:result}
\end{figure*} 

\section{Results \& Discussion}\label{results}

\begin{table*}[ht]
\centering
\label{tab:performance}
\caption{Comparison of pose estimation models on SMaL Dataset}
\begin{tabular}{|c|c|c|c|c|c|c|c|c|c|c|c|}
\hline
& & \multicolumn{3}{c|}{Pretraining} & \multicolumn{5}{c|}{Fold Accuracy} & \multicolumn{2}{c|}{Average}\\
\cline{3-12}    
Model & Params & Synth & Real & Epochs & 1 & 2 & 3 & 4 & 5 & PCK & NME\\
\hline     
HRNet W32 & 28.5M & - & - & - & 82.80 & 76.25 & 83.27 & 85.42 & 79.76 & 81.50 &	20.4 \\
HRNet-W48 & 63.6M & - & - & - & 84.17 & 80.00 & 85.42 & 87.74 & 83.63 & 84.19 &	18.6 \\
Transpose-R-A4 & 6.0M  & - & - & - & 77.92 & 73.87 & 78.87 & 81.85 & 78.33 & 78.17 & 23.2 \\
Transpose-H-A4 & 17.3M & - & - & - & 81.19 & 76.90 & 82.44 & 82.68 & 80.71 & 80.79 & 22.1 \\
Transpose-H-A6 & 17.5M & - & - & - & 81.96 & 79.35 & 83.99 & 83.99 & 80.83 & 82.02 & 20.7 \\
ViTPose-Base & 90.0M & - & - & - & 83.51 & 81.96 & 86.61 & 87.86 & 82.92 & 84.57 &	17.9 \\
ViTPose-S-100 & 90.0M & \checkmark & & 100 & 83.75 & 82.20 & 86.73 & 87.20 & 84.58 & 84.89 & 17.6 \\
ViTPose-R-100 & 90.0M & & \checkmark & 100 & 83.57 & 80.71 & 85.71 & 86.79 & 84.11 & 84.18 & 18.8 \\
ViTPose-B-100 & 90.0M & \checkmark & \checkmark & 100 & 84.76 & 82.44 & 86.67 & \textbf{89.64} & \textbf{85.89} & 85.88 & 17.2 \\
ViTPose-S-150 & 90.0M & \checkmark & & 150 & \textbf{85.77} & \textbf{83.21} & \textbf{87.92} & 88.75 & 84.64 & \textbf{86.06}* & \textbf{16.9}* \\
ViTPose-R-150 & 90.0M & & \checkmark & 150 & 83.57 & 81.49 & 85.83 & 87.44 & 84.17 & 84.50 & 18.3 \\
ViTPose-B-150 & 90.0M & \checkmark & \checkmark & 150 & 85.30 & 81.73 & 86.61 & 87.62 & 83.87 & 85.02 & 17.6 \\
\hline 
\end{tabular}
\end{table*}

\textbf{The best performing  pose estimation architecture on SMaL is ViTPose.} The performance of the pose estimation methods and their variants are shown in Table \ref{tab:performance}. From the models that were not pretrained, the best performing model was ViTPose, with a PCK of 84.57\% and a NME of 17.9 mm. The next top performing method was HRNet-W48, with a PCK of 84.19\% and a NME of 18.6 mm. Although the improvement in PCK marginal there was a significant increase in the NME, demonstrating that the ViTPose architecture has better localization than HRNet. The TransPose variants had the lowest overall performance of the baseline methods. The ResNet variant, TransPose-R-A4, had the lowest performance. TransPose-H-A4, which uses the HRNet-W32 backbone achieved a lower PCK than HRNet-W32 alone, similarly, the TransPose-H-A6 which uses the HRNet-W48 backbone achieved a lower PCK than the HRNet-W48 alone. Although TransPose was shown to achieve better performance on the COCO dataset when compared to the HRNet, this did not hold true when training on the SMaL dataset. This may be due to the fact that TransPose only uses 5\%-25\% of the original backbones, significantly reducing the knowledge transfer of the pretrained backbones. Given the relatively small dataset of the depth and PSM modalities, the advantage of larger pretrained backbones is evident. The larger variants of the baseline methods consistently outperform the the smaller models by significant margins. It is important to note, however, that although ViTPose has the largest number of parameters, it can achieve faster inference speeds than the other baseline methods \cite{vitpose}. 

\textbf{Pretraining the ViT backbone on depth and pressure imagery results in increased performance on SMaL.} There is also a clear benefit to pretraining the ViT backbone on data that is similar to the target dataset. Even though the pretraining data was solely adult data and the majority synthetic, it remained transferable to the SMaL dataset. Introducing a significant amount of unlabelled depth and PSM data to the model successfully reduced the domain shift between the original ImageNet training set and the target SMaL dataset. The best performing pretrained model was ViTPose-S-150 with a PCK of 86.06\% and a NME of 16.9 mm. The addition of 50 epochs of pretraining helped to increase performance over the ViT-S-100. This was the overall best performing model on both metrics and showed a statistically significant difference in performance when compared to ViT-Base ($p < 0.008$, where 0.05/6 since 6 pretraining variations were tested). It is important to note that when observing the overall performance across each fold, there is a natural variance between fold. Since each fold is only made of 60 unique poses, a few difficult cases within the fold can cause performance drops across all models. Therefore, to test for significance a paired t-test was used.  

Interestingly, using SLP real data alone for pretraining resulted in worse performance than no pretraining; both ViT-R-100 and ViT-R-150 performed worse than ViT-Base. This may be due to the relatively small dataset size of SLP, resulting in overfitting on the MAE task. Using both real and syntheic datasets for pretraining shown increased performance compared to synthetic only, as ViT-B-100 outperformed ViT-S-100 and was the second best pretraining strategy. ViT-B-100 is the equivalent to ViT-S-150 with continued training on the SLP dataset for another 100 epochs. Howevere, this was not the case for ViT-B-150, which had worse performance than ViT-S-150. This indicates that continued training on SLP for 100 epochs has a positive effect, but if this is increased by 50 epochs it has a negative effect. A potential improvement to the pretraining strategy could be training on synthetic data for 150 epochs and continuing training on real data for only 100 epochs, to avoid overfitting on the SLP data. Overall, these results show the benefit of hierarchical pretraining for under the cover pose estimation as long as there is a large enough dataset and caution is used to avoid overfitting when pretraining on smaller datasets. 
\clearpage
\textbf{Under the cover infant pose can be accurately inferred using a pose estimation model.}
Figure \ref{fig:result} shows various test cases from SMaL and the best performing method, ViTPose-S-150's predictions. The accuracy of the pose predictions are impressive given the heavy occlusions present in the depth image and the low resolution of the pressure image. When compared to the performance on other datasets, our performance is comparable. Wu \textit{et al.} achieve a root-mean-square distance on all joints of 4.12 pixels on their uncovered depth based infant dataset \cite{Wu2022}. When converting our best performance of 16.9mm to pixels, we achieve an error of 3.33 pixels. The model gets the majority of keypoints correct with most of the error on joints in the extremity. The slight errors on the on wrist and ankle are observed in the first and third row of Figure \ref{fig:result}. When breaking down the performance by joint type, as shown in Table \ref{tab:jointacc} and \ref{tab:jointnme}, it is clear that the wrist poses the most difficult challenge to the model, with the highest NME error and the lowest PCK. The hip and shoulder keypoints boast the highest accuracy of the joints. This is relevant to one of our motivations, under the cover ROI detection, as the bounding box connecting these joints highlights the most important regions for noncontact RR estimation \cite{Maurya2021}. Table \ref{tab:jointacc} and \ref{tab:jointnme} also show the added difficulty when using thicker blanket covering, as the overall performance was worse under thick blankets, although not by a large margin. The third and fourth rows of Figure \ref{fig:result} show an example of the thick cover condition. The fourth row presents an additional challenge, with the pose being very tight and the limbs close together. In spite of this, the majority of the prediction was correct. 

\begin{table}[ht]
\centering
\caption{Per-joint accuracy of ViTPose-S-150 for both cover conditions}
\label{tab:jointacc}
\begin{tabu}{|c|c|c|[1pt]c|}
\hline
\textbf{Joint} &  \textbf{Thin Cover} & \textbf{Thick Cover}& \textbf{Average}\\
\hline         
R-Ankle & 81.67 & 83.67 & 82.67\\
L-Ankle	& 84.67	& 82.00	& 83.33\\
\hline
R-Knee	& 88.67	& 86.33	& 87.50\\
L-Knee	& 87.00	& 86.67	& 86.83\\
\hline
R-Hip	& 92.00	& 92.33	& 92.17\\
L-Hip	& 92.33	& 92.33	& 92.33\\
\hline
R-Wrist	& 77.00	& 71.67	& 74.33\\
L-Wrist	& 77.67	& 78.00	& 77.83\\
\hline
R-Elbow	& 89.33	& 82.67	& 86.00\\
L-Elbow	& 90.33	& 85.67	& 88.00\\
\hline
R-Shoulder	& 92.67	& 90.33	& 91.50\\
L-Shoulder	& 91.00	& 90.67	& 90.83\\
\hline
Thorax	& 97.00	& 95.67	& 96.33\\
Head	& 74.67	& 75.67	& 75.17\\
\tabucline[1pt]{-}
\textbf{Average}& 86.86 & 85.26 & 86.06\\
\hline
\end{tabu}
\end{table}

\begin{table}[ht]
\centering
\caption{Per-joint NME (mm) of ViTPose-S-150 for both cover conditions}
\label{tab:jointnme}
\begin{tabu}{|c|c|c|[1pt]c|}
\hline
\textbf{Joint} &  \textbf{Thin Cover} & \textbf{Thick Cover}& \textbf{Average}\\
\hline         
R-Ankle & 19.4 & 18.7 & 19.1\\
L-Ankle	& 19.0 & 19.7 & 19.4\\
\hline
R-Knee	& 16.3 & 16.6 & 16.4\\
L-Knee	& 16.5 & 17.6 & 17.1\\
\hline
R-Hip   & 12.3 & 12.2 & 12.2\\
L-Hip   & 12.0 & 12.2 & 12.1\\
\hline
R-Wrist	& 26.3 & 28.8 & 27.5\\
L-Wrist	& 22.6 & 24.0 & 23.3\\
\hline
R-Elbow	& 17.6 & 21.0 & 19.3\\
L-Elbow	& 15.7 & 17.7 & 16.7\\
\hline
R-Shoulder	& 11.7	& 12.7	& 12.2\\
L-Shoulder	& 13.0	& 13.0	& 13.0\\
\hline
Thorax	& 10.7	& 11.1	& 10.9\\
Head	& 17.5	& 18.3	& 17.9\\
\tabucline[1pt]{-}
\textbf{Average}& 16.5	& 17.4	& 16.9\\
\hline
\end{tabu}
\end{table}

\textbf{There is significant advantage to transferring knowledge from adult pose to infant pose.}
The two stage training strategy is highly effective to improve performance on the SMaL dataset, shown in Table \ref{tab:ablation}. When excluding the SLP training stage, the performance drops 12\% and 9\% for ViTPose-Base and ViTPose-S-150 respectively. This finding supports the conclusions of our previous work where adult pressure data was leveraged for transfer learning to infant pressure data \cite{Kyrollos2022}. The benefit of transferring knowledge from adult data is still apparent even when no labelled data is used. This is seen from the large performance gains after pretraining on unlabelled adult data, with a 9\% increase of ViTPose-S-150 over ViTPose-Base after only training on SMaL data. With the abundance of available adult data and the relative lack of infant data, this strategy can be applied to any task involving the domain shift between adults and infants.

\textbf{Both depth and PSM inputs contribute to model performance.}
When limiting the modality input to only depth or only PSM, we see significant drops in performance, as shown in Figure \ref{tab:ablation}. The combination of these modalities contributes more than a 10\% increase in performance, pointing to the completely nature of the depth and pressure images. In all cases, the PSM only training outperforms the depth only, showing that individually the PSM is more informative than depth for the SMaL dataset. When the models are not trained on SLP data first (also using a single modality), we see larger performance drops when using depth alone compared to PSM alone. In addition, the disparity between the modalities is exaggerated when no training is performed on SLP. This shows that there is higher variance in the depth data compared to the PSM data, however this can be mitigated by first training on a larger depth only dataset before. 

\begin{table}[ht]
\centering
\label{tab:ablation}
\caption{Effects of modality and training data on PCK}
\begin{tabu}{|c|c|c|c|c|c|c|}
\hline
& \multicolumn{3}{c|}{\textbf{SLP + SMaL}} & \multicolumn{3}{c|}{\textbf{SMaL Only}}\\
Model & Both & Depth & PSM & Both & Depth & PSM\\
\hline 
ViTPose-Base  & 84.57 & 71.30 & 74.57 & 72.48 & 54.27 & \textbf{66.49}\\
\hline 
ViTPose-S-150 & \textbf{86.06} & \textbf{73.56} & \textbf{75.02} & \textbf{77.95} & \textbf{57.02} & 64.81\\
\hline
\end{tabu}
\end{table}

\section{Conclusion}
In summary, we present and release a novel multimodal dataset, SMaL, for under the cover infant pose estimation. We use this dataset to train state-of-the-art pose estimation methods and evaluate their performance. We also leverage existing multimodal adult pose datasets for transfer learning. By implementing a transformer-based pose estimation method, we demonstrate a hierarchical pretraining and finetuning strategy using adult data to significantly boost performance on our SMaL dataset. Overall, we were able to accurately infer infant pose in the presence of heavy occlusion from blankets using depth and pressure imagery.
\clearpage
\bibliography{references.bib}

\begin{thebibliography}{10}
\providecommand{\url}[1]{#1}
\csname url@samestyle\endcsname
\providecommand{\newblock}{\relax}
\providecommand{\bibinfo}[2]{#2}
\providecommand{\BIBentrySTDinterwordspacing}{\spaceskip=0pt\relax}
\providecommand{\BIBentryALTinterwordstretchfactor}{4}
\providecommand{\BIBentryALTinterwordspacing}{\spaceskip=\fontdimen2\font plus
\BIBentryALTinterwordstretchfactor\fontdimen3\font minus
  \fontdimen4\font\relax}
\providecommand{\BIBforeignlanguage}[2]{{%
\expandafter\ifx\csname l@#1\endcsname\relax
\typeout{** WARNING: IEEEtran.bst: No hyphenation pattern has been}%
\typeout{** loaded for the language `#1'. Using the pattern for}%
\typeout{** the default language instead.}%
\else
\language=\csname l@#1\endcsname
\fi
#2}}
\providecommand{\BIBdecl}{\relax}
\BIBdecl

\bibitem{Hamilton2000}
P.~Hamilton, M.~Curley, and R.~Aimi, ``Effect of adaptive motion-artifact
  reduction on qrs detection,'' \emph{Biomedical instrumentation \&
  technology}, vol.~34, no.~3, p. 197—202, 2000.

\bibitem{Wolf1996}
M.~Wolf, M.~Keel, K.~von Siebenthal, H.~Bucher, K.~Geering, Y.~Lehareinger, and
  P.~Niederer, ``Improved monitoring of preterm infants by fuzzy logic,''
  \emph{Technology and health care : official journal of the European Society
  for Engineering and Medicine}, vol.~4, no.~2, p. 193—201, August 1996.

\bibitem{adhesive1997}
C.~H. Lund, L.~B. Nonato, J.~M. Kuller, L.~S. Franck, C.~Cullander, and D.~K.
  Durand, ``Disruption of barrier function in neonatal skin associated with
  adhesive removal,'' \emph{The Journal of Pediatrics}, vol. 131, no.~3, pp.
  367 -- 372, 1997.

\bibitem{Kyrollos2021-saspsm}
D.~G. Kyrollos, K.~Greenwood, J.~Harrold, and J.~R. Green, ``Detection of false
  alarms in the {NICU} using pressure sensitive mat,'' in \emph{2021 {IEEE}
  Sensors Applications Symposium ({SAS})}.\hskip 1em plus 0.5em minus
  0.4em\relax {IEEE}, Aug. 2021.

\bibitem{Maurya2021}
L.~Maurya, P.~Kaur, D.~Chawla, and P.~Mahapatra, ``Non-contact breathing rate
  monitoring in newborns: A review,'' \emph{Computers in Biology and Medicine},
  vol. 132, p. 104321, May 2021.

\bibitem{Kyrollos2021-sasrr}
D.~G. Kyrollos, J.~B. Tanner, K.~Greenwood, J.~Harrold, and J.~R. Green,
  ``Noncontact neonatal respiration rate estimation using machine vision,'' in
  \emph{2021 {IEEE} Sensors Applications Symposium ({SAS})}.\hskip 1em plus
  0.5em minus 0.4em\relax {IEEE}, Aug. 2021.

\bibitem{Nizami2018}
S.~Nizami, A.~Bekele, M.~Hozayen, K.~J. Greenwood, J.~Harrold, and J.~R. Green,
  ``Measuring uncertainty during respiratory rate estimation using
  pressure-sensitive mats,'' \emph{{IEEE} Transactions on Instrumentation and
  Measurement}, vol.~67, no.~7, pp. 1535--1542, Jul. 2018.

\bibitem{aap1992positioning}
A.~T.~F. on~Infant~Positioning and SIDS, ``Positioning and sids,''
  \emph{Pediatrics}, vol.~89, no.~6, pp. 1120--1126, 1992.

\bibitem{Bembich2012}
S.~Bembich, C.~Oretti, L.~Travan, A.~Clarici, S.~Massaccesi, and S.~Demarini,
  ``Effects of prone and supine position on cerebral blood flow in preterm
  infants,'' \emph{The Journal of Pediatrics}, vol. 160, no.~1, pp. 162--164,
  Jan. 2012.

\bibitem{Fister2020}
P.~Fister, M.~Nolimal, H.~Lenasi, and M.~Klemenc, ``The effect of sleeping
  position on heart rate variability in newborns,'' \emph{{BMC} Pediatrics},
  vol.~20, no.~1, Apr. 2020.

\bibitem{Kukkola2022}
H.-L. Kukkola and T.~Kirjavainen, ``Obstructive sleep apnea is position
  dependent in young infants,'' \emph{Pediatric Research}, Aug. 2022.

\bibitem{SLPLiu2022}
S.~Liu, X.~Huang, N.~Fu, C.~Li, Z.~Su, and S.~Ostadabbas,
  ``Simultaneously-collected multimodal lying pose dataset: Enabling in-bed
  human pose monitoring,'' \emph{{IEEE} Transactions on Pattern Analysis and
  Machine Intelligence}, pp. 1--1, 2022.

\bibitem{Huang2021}
X.~Huang, N.~Fu, S.~Liu, and S.~Ostadabbas, ``Invariant representation learning
  for infant pose estimation with small data,'' in \emph{2021 16th {IEEE}
  International Conference on Automatic Face and Gesture Recognition ({FG}
  2021)}.\hskip 1em plus 0.5em minus 0.4em\relax {IEEE}, Dec. 2021.

\bibitem{Clever2022}
H.~M.~M. Clever, P.~Grady, G.~Turk, and C.~C. Kemp, ``{BodyPressure} -
  inferring body pose and contact pressure from a depth image,'' \emph{{IEEE}
  Transactions on Pattern Analysis and Machine Intelligence}, pp. 1--1, 2022.

\bibitem{reed2022self}
C.~J. Reed, X.~Yue, A.~Nrusimha, S.~Ebrahimi, V.~Vijaykumar, R.~Mao, B.~Li,
  S.~Zhang, D.~Guillory, S.~Metzger \emph{et~al.}, ``Self-supervised
  pretraining improves self-supervised pretraining,'' in \emph{Proceedings of
  the IEEE/CVF Winter Conference on Applications of Computer Vision}, 2022, pp.
  2584--2594.

\bibitem{Kyrollos2022}
D.~G. Kyrollos, K.~Greenwood, J.~Harrold, and J.~R. Green, ``Transfer learning
  approaches for neonate head localization from pressure images,'' in
  \emph{2022 {IEEE} International Symposium on Medical Measurements and
  Applications ({MeMeA})}.\hskip 1em plus 0.5em minus 0.4em\relax {IEEE}, Jun.
  2022.

\bibitem{Liu2017}
S.~Liu and S.~Ostadabbas, ``A vision-based system for in-bed posture
  tracking,'' in \emph{2017 {IEEE} International Conference on Computer Vision
  Workshops ({ICCVW})}.\hskip 1em plus 0.5em minus 0.4em\relax {IEEE}, Oct.
  2017.

\bibitem{Davoodnia2021}
V.~Davoodnia, S.~Ghorbani, and A.~Etemad, ``In-bed pressure-based pose
  estimation using image space representation learning,'' in \emph{{ICASSP}
  2021 - 2021 {IEEE} International Conference on Acoustics, Speech and Signal
  Processing ({ICASSP})}.\hskip 1em plus 0.5em minus 0.4em\relax {IEEE}, Jun.
  2021.

\bibitem{Clever2018}
H.~M. Clever, A.~Kapusta, D.~Park, Z.~Erickson, Y.~Chitalia, and C.~C. Kemp,
  ``3d human pose estimation on a configurable bed from a pressure image,'' in
  \emph{2018 {IEEE}/{RSJ} International Conference on Intelligent Robots and
  Systems ({IROS})}.\hskip 1em plus 0.5em minus 0.4em\relax {IEEE}, Oct. 2018.

\bibitem{Casas2018}
L.~Casas, N.~Navab, and S.~Demirci, ``Patient 3d body pose estimation from
  pressure imaging,'' \emph{International Journal of Computer Assisted
  Radiology and Surgery}, vol.~14, no.~3, pp. 517--524, Dec. 2018.

\bibitem{Clever2020}
H.~M. Clever, Z.~Erickson, A.~Kapusta, G.~Turk, C.~K. Liu, and C.~C. Kemp,
  ``Bodies at rest: 3d human pose and shape estimation from a pressure image
  using synthetic data,'' in \emph{2020 {IEEE}/{CVF} Conference on Computer
  Vision and Pattern Recognition ({CVPR})}.\hskip 1em plus 0.5em minus
  0.4em\relax {IEEE}, Jun. 2020.

\bibitem{Kyrollos2021-i2mtc}
D.~G. Kyrollos, R.~Hassan, Y.~S. Dosso, and J.~R. Green, ``Fusing
  pressure-sensitive mat data with video through multi-modal registration,'' in
  \emph{2021 {IEEE} International Instrumentation and Measurement Technology
  Conference (I2MTC)}.\hskip 1em plus 0.5em minus 0.4em\relax {IEEE}, May 2021.

\bibitem{Cao2022human}
T.~Cao, M.~A. Armin, S.~Denman, L.~Petersson, and D.~Ahmedt-Aristizabal,
  ``In-bed human pose estimation from unseen and privacy-preserving image
  domains,'' in \emph{2022 {IEEE} 19th International Symposium on Biomedical
  Imaging ({ISBI})}.\hskip 1em plus 0.5em minus 0.4em\relax {IEEE}, Mar. 2022.

\bibitem{Dayarathna2022}
T.~Dayarathna, T.~Muthukumarana, Y.~Rathnayaka, S.~Denman, C.~de~Silva,
  A.~Pemasiri, and D.~Ahmedt-Aristizabal, ``Privacy-preserving in-bed pose
  monitoring: A fusion and reconstruction study,'' 2022.

\bibitem{Yin2022}
Y.~Yin, J.~P. Robinson, and Y.~Fu, ``Multimodal in-bed pose and shape
  estimation under the blankets,'' 2020.

\bibitem{Afham2022}
M.~Afham, U.~Haputhanthri, J.~Pradeepkumar, M.~Anandakumar, A.~D. Silva, and
  C.~U.~S. Edussooriya, ``Towards accurate cross-domain in-bed human pose
  estimation,'' in \emph{{ICASSP} 2022 - 2022 {IEEE} International Conference
  on Acoustics, Speech and Signal Processing ({ICASSP})}.\hskip 1em plus 0.5em
  minus 0.4em\relax {IEEE}, May 2022.

\bibitem{Groos2022}
D.~Groos, L.~Adde, R.~St{\o}en, H.~Ramampiaro, and E.~A. Ihlen, ``Towards
  human-level performance on automatic pose estimation of infant spontaneous
  movements,'' \emph{Computerized Medical Imaging and Graphics}, vol.~95, p.
  102012, Jan. 2022.

\bibitem{Cao2022}
X.~Cao, X.~Li, L.~Ma, Y.~Huang, X.~Feng, Z.~Chen, H.~Zeng, and J.~Cao,
  ``{AggPose}: Deep aggregation vision transformer for infant pose
  estimation,'' in \emph{Proceedings of the Thirty-First International Joint
  Conference on Artificial Intelligence}.\hskip 1em plus 0.5em minus
  0.4em\relax International Joint Conferences on Artificial Intelligence
  Organization, Jul. 2022.

\bibitem{Hesse2019}
N.~Hesse, C.~Bodensteiner, M.~Arens, U.~G. Hofmann, R.~Weinberger, and A.~S.
  Schroeder, ``Computer vision for medical infant motion analysis: State of the
  art and {RGB}-d data set,'' in \emph{Lecture Notes in Computer
  Science}.\hskip 1em plus 0.5em minus 0.4em\relax Springer International
  Publishing, 2019, pp. 32--49.

\bibitem{Moccia2020}
S.~Moccia, L.~Migliorelli, V.~Carnielli, and E.~Frontoni, ``Preterm infants'
  pose estimation with spatio-temporal features,'' \emph{{IEEE} Transactions on
  Biomedical Engineering}, vol.~67, no.~8, pp. 2370--2380, Aug. 2020.

\bibitem{Donati2013}
M.~Donati, F.~Cecchi, F.~Bonaccorso, M.~Branciforte, P.~Dario, and N.~Vitiello,
  ``A modular sensorized mat for monitoring infant posture,'' \emph{Sensors},
  vol.~14, no.~1, pp. 510--531, Dec. 2013.

\bibitem{Rihar2014}
A.~Rihar, M.~Mihelj, J.~Pa{\v{s}}i{\v{c}}, J.~Kolar, and M.~Munih, ``Infant
  trunk posture and arm movement assessment using pressure mattress, inertial
  and magnetic measurement units ({IMUs}),'' \emph{Journal of
  {NeuroEngineering} and Rehabilitation}, vol.~11, no.~1, p. 133, 2014.

\bibitem{Wu2022}
Q.~Wu, G.~Xu, F.~Wei, J.~Kuang, P.~Qin, Z.~Li, and S.~Zhang, ``Supine infant
  pose estimation via single depth image,'' \emph{{IEEE} Transactions on
  Instrumentation and Measurement}, vol.~71, pp. 1--11, 2022.

\bibitem{he2022masked}
K.~He, X.~Chen, S.~Xie, Y.~Li, P.~Doll{\'a}r, and R.~Girshick, ``Masked
  autoencoders are scalable vision learners,'' in \emph{Proceedings of the
  IEEE/CVF Conference on Computer Vision and Pattern Recognition}, 2022, pp.
  16\,000--16\,009.

\bibitem{vaswani2017attention}
A.~Vaswani, N.~Shazeer, N.~Parmar, J.~Uszkoreit, L.~Jones, A.~N. Gomez,
  {\L}.~Kaiser, and I.~Polosukhin, ``Attention is all you need,''
  \emph{Advances in neural information processing systems}, vol.~30, 2017.

\bibitem{dosovitskiy2020image}
A.~Dosovitskiy, L.~Beyer, A.~Kolesnikov, D.~Weissenborn, X.~Zhai,
  T.~Unterthiner, M.~Dehghani, M.~Minderer, G.~Heigold, S.~Gelly, J.~Uszkoreit,
  and N.~Houlsby, ``An image is worth 16x16 words: Transformers for image
  recognition at scale,'' in \emph{International Conference on Learning
  Representations}, 2021.

\bibitem{peng2022beit}
Z.~Peng, L.~Dong, H.~Bao, Q.~Ye, and F.~Wei, ``Beit v2: Masked image modeling
  with vector-quantized visual tokenizers,'' \emph{arXiv preprint
  arXiv:2208.06366}, 2022.

\bibitem{li2022exploring}
Y.~Li, H.~Mao, R.~Girshick, and K.~He, ``Exploring plain vision transformer
  backbones for object detection,'' \emph{arXiv preprint arXiv:2203.16527},
  2022.

\bibitem{vitpose}
Y.~Xu, J.~Zhang, Q.~Zhang, and D.~Tao, ``Vitpose: Simple vision transformer
  baselines for human pose estimation,'' 2022.

\bibitem{abnar2021exploring}
S.~Abnar, M.~Dehghani, B.~Neyshabur, and H.~Sedghi, ``Exploring the limits of
  large scale pre-training,'' in \emph{International Conference on Learning
  Representations}, 2022.

\bibitem{hrnet}
K.~Sun, B.~Xiao, D.~Liu, and J.~Wang, ``Deep high-resolution representation
  learning for human pose estimation,'' in \emph{Proceedings of the IEEE/CVF
  conference on computer vision and pattern recognition}, 2019, pp. 5693--5703.

\bibitem{transpose}
S.~Yang, Z.~Quan, M.~Nie, and W.~Yang, ``Transpose: Keypoint localization via
  transformer,'' 2020.

\end{thebibliography}
\bibliographystyle{IEEEtran}
\end{document}